# An Indirect Genetic Algorithm for a Nurse Scheduling Problem

Uwe Aickelin
School of Computer Science
University of Nottingham
NG8 1BB   UK
uxa@cs.nott.ac.uk

Kathryn A. Dowsland
Gower Optimal Algorithms Ltd.
5, Whitestone Lane
Newton
Swansea
SA3 4UH
UK

Correspondence to U Aickelin

**Abstract**
This paper describes a Genetic Algorithms approach to a manpower-scheduling problem arising at a major UK hospital. Although Genetic Algorithms have been successfully used for similar problems in the past, they always had to overcome the limitations of the classical Genetic Algorithms paradigm in handling the conflict between objectives and constraints. The approach taken here is to use an indirect coding based on permutations of the nurses, and a heuristic decoder that builds schedules from these permutations. Computational experiments based on 52 weeks of live data are used to evaluate three different decoders with varying levels of intelligence, and four well-known crossover operators. Results are further enhanced by introducing a hybrid crossover operator and by making use of simple bounds to reduce the size of the solution space. The results reveal that the proposed algorithm is able to find high quality solutions and is both faster and more flexible than a recently published Tabu Search approach.

An Indirect Genetic Algorithm for a Nurse Scheduling Problem

# 1 The Nurse Scheduling Problem

In recent years, Genetic Algorithms (GAs) have become increasingly popular for solving complex optimisation problems such as those found in the areas of scheduling or timetabling. Unfortunately, there is no pre-defined way of including constraints into GAs. This is probably one of their biggest drawbacks, as it does not make them readily amenable to most real world optimisation problems. Some methods for dealing with constraints do exist, notably penalty and repair functions. However, as noted by Michalewicz [1], their application and success is problem specific. Here we use an alternative approach: the genetic operators act on an unconstrained solution space whose elements are converted into solutions by a schedule builder that acts as the decoder for the GA. Our work differs from previous studies using a similar strategy in two ways. First, constraints are used actively to guide the heuristic reducing the reliance on penalty functions. However, the nature of our problem means that it is not possible for the schedule-builder to guarantee feasible solutions and much of our investigation involves finding the best way of balancing the conflicting issues of feasibility and solution cost. Second, we investigate the effectiveness of various different permutation crossover operators for implementations of this type, and consequently introduce a hybrid crossover that combines the desirable features of the best two.

The work described in this paper has two objectives. The first is to develop a fast, flexible solution approach to a nurse rostering problem and second, to add to the body of knowledge on solving constrained problems using GAs. Although there are many published algorithms for nurse scheduling, some characteristics of our particular problem, as outlined in the next section, make these approaches unsuitable for us. An extensive summary of these methods can be found in Hung [2], Sitompul and Randhawa [3] and Bradley and Martin [4]. Examples of approaches based on GAs include Easton and Mansour [5] and Tanomaru [6]. Those who have tackled the problem considered here have met with varying degrees of success. For example, Fuller [7] uses an integer programming approach. She showed that in order to solve the problem reliably within a reasonable amount of computational time it is necessary to use a variety of features that are only available in specialist IP packages. Without these features, around one third of instances required several hours of run-time to find good feasible solutions. Due to the high costs of such software, this is not a practical solution for many hospitals. Dowsland [8] solves the problem using a Tabu Search approach. Her results are impressive in solving instances arising at a particular hospital, but are less good on data with different characteristics.

The approach adopted here was motivated by previous research into a GA solution to the problem. Aickelin and Dowsland [9] show that a straightforward GA implementation is not able to tackle the problem. The addition of problem specific information, in the form of co-evolving sub-populations and a modified fitness function coupled with an intelligent hill-climber, yields reasonable solutions in a very short time. However, solution quality is not consistent and the nature of the modifications means that the approach may not be robust to small changes in problem specification. Thus, there is still room for improvement.

Here we use a different strategy for a GA approach, in which the individuals in the population do not represent direct encodings of solutions. Instead, solutions are obtained via separate decoder heuristics that build solutions from permutations of the list of available nurses using the constraints as guides. These permutations are manipulated by the GA. The advantage of this strategy is that the GA can remain canonical, i.e. it solves an unconstrained problem and does not require a hill-climber or other problem-specific knowledge. Note that penalty functions might still be required if the decoder fails to find a feasible solution. This also makes the algorithm more flexible than the integrated approaches outlined above, as the GA component can remain unchanged if the problem changes, e.g. if there are new constraints or objectives.

The idea of an indirect GA is not new, and similar approaches have been used for other problems, for example by Davis [10] and Palmer and Kershenbaum [11]. However, there is one major difference between our work and theirs: The nurse scheduling problem is not a simple permutation problem like job shop scheduling but has additional constraints and hence our decoders cannot guarantee to always produce feasible solutions. Thus, there is still some work to be done by a penalty function. Our computational experiments are designed to examine this aspect of the implementation, as well as investigating the influence of the choice of heuristic decoder and the crossover operator on solution speed and quality. The result is an algorithm that is still simple but robust enough to cope with this infeasibility. This is achieved by striking a balance between the stochastic elements of the GA and the deterministic nature of the proposed decoder. Our key tools for this are the setting of parameter values, deciding on genetic strategies and finding a suitable decoder function.

## 2  The Problem

The rostering problem tackled in this paper can be described as follows. The task is to create weekly schedules for wards of up to 30 nurses by assigning one of a number of possible shift patterns to each nurse. These schedules have to satisfy working contracts and meet the demand for a given number of nurses of different grades on each shift, while being seen to be fair by the staff concerned. The latter objective is achieved by meeting as many of the nurses' requests as possible and considering historical information to ensure that unsatisfied requests and unpopular shifts are evenly distributed. The problem is complicated by the fact that higher qualified nurses can substitute less qualified nurses but not vice versa. Thus scheduling the different grades independently is not possible. Furthermore, the problem has a special day-night structure as most of the nurses are contracted to work either days or nights in one week but not both. However due to working contracts, the number of days worked is not usually the same as the number of nights. Therefore, it becomes important to schedule the 'right' nurses onto days and nights respectively. It is the latter two characteristics that make this problem challenging for any local search algorithm as finding and maintaining feasible solutions is extremely difficult. Furthermore, due to this special structure previous nurse scheduling algorithms suggested in the literature cannot be used. Further details of the nurse-scheduling problem can be found in Aickelin and Dowsland [9].

As described in Dowsland and Thompson [12] the problem can be decomposed into three independent stages. The first stage ensures that there are enough nurses to provide adequate cover. The second stage assigns the nurses to the correct number of day or night shifts. A final phase allocates those working on particular day to the early or late shift on that day. Phases 1 and 3 are easily solved using classical optimisation models. Thus, this paper deals with the highly constrained second step.

The numbers of days or nights to be worked by each nurse defines the set of feasible weekly work patterns for that nurse. These will be referred to as shift patterns or shift pattern vectors in the following. For each nurse $i$ and each shift pattern $j$ all the information concerning the desirability of the pattern for this nurse is captured in a single numeric preference cost $p_{ij}$. This was done in close consultation with the hospital and is a weighted sum of the following factors: Basic shift-pattern cost, general day / night preferences, specific requests, continuity problems, number of successive working day, rotating nights / weekends and other working history information.

Patterns that violate mandatory contractual requirements are marked as infeasible for a particular nurse and week. Again, further details can be found in Dowsland and Thompson [12]

The problem can be formulated as an integer linear program as follows.

Indices:

$i = 1...n$ nurse index.

$j = 1...m$ shift pattern index.

$k = 1...14$ day and night index (1...7 are days and 8...14 are nights).

$s = 1...p$ grade index.

Decision variables:

$$x_{ij} = \begin{cases} 1 & \text{nurse } i \text{ works shift pattern } j \\ 0 & \text{else} \end{cases}$$

Parameters:

$n$ = Number of nurses.

$m$ = Number of shift patterns.

$p$ = Number of grades.

$$a_{jk} = \begin{cases} 1 & \text{shift pattern } j \text{ covers day / night } k \\ 0 & \text{else} \end{cases}$$

$$q_{is} = \begin{cases} 1 & \text{nurse } i \text{ is of grade } s \text{ or higher} \\ 0 & \text{else} \end{cases}$$

$p_{ij}$ = Preference cost of nurse $i$ working shift pattern $j$.

$N_i$ = Working shifts per week of nurse $i$ if night shifts are worked.

$D_i$ = Working shifts per week of nurse $i$ if day shifts are worked.

$B_i$ = Working shifts per week of nurse $i$ if both day and night shifts are worked.

$R_{ks}$ = Demand of nurses with grade $s$ on day / night $k$.

$F(i)$ = Set of feasible shift patterns for nurse $i$, where $F(i)$ is defined as

$$F(i) = \begin{cases} \sum_{k=1}^{7} a_{jk} = D_i & \forall j \in \text{day shifts} \\ \text{or} \\ \sum_{k=8}^{14} a_{jk} = N_i & \forall j \in \text{night shifts} \\ \text{or} \\ \sum_{k=1}^{14} a_{jk} = B_i & \forall j \in \text{combined shifts} \end{cases} \quad \forall i$$

Target function: Minimise total preference cost of all nurses

$$\sum_{i=1}^{n} \sum_{j \in F(i)}^{m} p_{ij} x_{ij} \quad \rightarrow \quad \text{min!}$$

Subject to:

1. Every nurse works exactly one feasible shift pattern:

$$\sum_{j \in F(i)} x_{ij} = 1 \qquad \forall i \qquad (1)$$

2. The demand for nurses is fulfilled for every grade on every day and night:

$$\sum_{j \in F(i)} \sum_{i=1}^{n} q_{is} a_{jk} x_{ij} \geq R_{ks} \qquad \forall k, s \qquad (2)$$

Constraint set (1) ensures that every nurse works exactly one shift pattern from his/her feasible set, and constraint set (2) ensures that the demand for nurses is covered for every grade on every day and night. Note that the definition of $q_{is}$ is such that higher graded nurses can substitute those at lower grades if necessary. Typical problem dimensions are 30 nurses of three grades and 411 shift patterns. Thus, the IP formulation has about 12000 binary variables and 100 constraints. Although this is only a moderately sized problem, Fuller [7] shows that some problem instances remain unsolved after hours of computation time on a Pentium II PC (equivalent to the hospital's hardware) using professional software.

## 3 Genetic Algorithms

GAs are generally attributed to Holland [13] and his students in the 1970s, although evolutionary computation dates back further (refer to Fogel [14] for an extensive review of early approaches). GAs are stochastic meta-heuristics that mimic some features of natural evolution. Canonical GAs were not intended for function optimisation, as discussed by De Jong [15]. However, slightly modified versions proved very successful. For an introduction to GAs for function optimisation, see Deb [16]. Many examples of successful implementations can be found in Bäck [17], Chaiyaratana and Zalzala [18] and others.

In a nutshell, GAs mimic the evolutionary process and the idea of the survival of the fittest. Starting with a population of randomly created solutions, better ones are more likely to be chosen for recombination into new solutions, i.e. the fitter a solution, the more likely it is to pass on its information to future generations of solutions. In addition to recombining solutions, new solutions may be formed through mutating or randomly changing old solutions. Some of the best solutions of each generation are kept whilst the others are replaced by the newly formed solutions. The process is repeated until stopping criteria are met.

However, constrained optimisation with GAs remains difficult. The root of the problem is that simply following the building block hypothesis, i.e. combining good building blocks or partial solutions to form good full solutions, is no longer enough, as this does not check for constraint consistency. To solve this dilemma, many ideas have been proposed of which the major ones (penalty and repair functions) will be briefly outlined in the following. A good overview of these and most other techniques can be found in Michalewicz [1].

Penalty functions try to avoid infeasible solutions by steering the search away from them, whilst repair functions try to 'fix' such solutions so that they become feasible. Unfortunately, penalising infeasible solutions is no guarantee that the GA will succeed in finding good feasible solutions, as a highly fit but slightly infeasible solution might dominate the search. If the solution space is dominated by infeasible solutions then finding good feasible solutions with penalty functions alone is unlikely to be successful. Using more complex penalty functions, such as dynamic or adaptive schemes can help, but traditionally most such approaches provide a means of comparing infeasible and feasible solutions during the selection stages rather than dispensing with the problem of satisfying constraints. More promising seem approaches such as used by Beasley and Chu [19] that bias the way the solution space is sampled towards feasible areas. We will use such a scheme in our decoder.

Repairing infeasible solutions also has its drawbacks. Firstly, it is often as difficult to repair an invalid solution as it is to find a good feasible solution. Secondly, repeated repair might lead to a build up of poor material within the population, as there is not enough incentive for development. Finally, repair routines are typically time consuming and it is arguable that this time is better spent on a more direct search of the solution space. Furthermore, the implementation of both penalty and repair functions is highly problem specific and successful applications cannot usually be transferred to other problems. In our case, it is not possible to find a simple, fast repair mechanism that will convert an arbitrary infeasible solution into a feasible one.

The approach presented here is the combination of an indirect GA with a separate heuristic decoder function. The GA tries to find the best possible ordering of the nurses, which is then fed into a greedy decoder that builds the actual solution. One way of looking at this decoder is as an extended fitness function calculation, i.e. the decoder determines the fitness of a solution once it has built a schedule from the permutation of nurses. One advantage of this approach is that all problem specific information is contained within the decoder, whilst the GA can be kept canonical. The only difference from a standard GA is the need for permutation-based crossover and mutation operators as explained for instance in Goldberg [20]. This should allow for easy adaptation to different problems. Furthermore, we will show that such a decoder allows us to use constraints more actively in addition to a background penalty function. These issues will be investigated.

There are similar approaches, reported in the literature, which use decoders that always assemble feasible solutions. This is possible due to the characteristics of the problems studied. Examples are Davis [10] and Fang et al. [21] for job shop scheduling and Podgorelec and Kokol [22] and Corne and Odgen [23] for timetabling problems. In essence, there the decoder is only a schedule builder, whereas in our case it is also a constraint handler that biases the search towards the feasible region. As there is no obvious way of constructing a decoder that only assembles feasible solutions, this research extends previous studies using indirect decoders by allowing infeasible solutions in the solution space. This raises the question as to how the balance between quality and feasibility should be handled in the decoder. We investigate this issue and show that a combination of a stochastic GA and a deterministic and balanced decoder successfully solves the nurse scheduling problem by finding the optimal solution in more than half of all problem instances tackled and being on average within 2% of optimality.

It is also worth noting that our approach does not follow all of the rules suggested by Palmer and Kershenbaum [11] for decoders. Their rules are:

1) For each solution in the original space, there is a solution in the encoded space.
2) Each encoded solution corresponds to one feasible solution in the original space.
3) All solutions in the original space should be represented by the same number of encoded solutions.
4) The transformation between solutions should be computationally fast.
5) Small changes in the encoded solution should result in small changes in the solution itself.

The idea here is that the new solution space should not introduce bias by excluding some solutions, or by representing some by more points than others. In our case rules 1) and 3) are violated, but we argue that such violation is a desirable feature of an indirect representation. The use of a greedy decoder will mean that some solutions are not represented at all. However, these will tend to be lower quality solutions. Similarly some solutions will be represented by more than one permutation, but as these are likely to be feasible or of low cost, the net result is a biased search spending more time in high quality areas of the solution space. However, as our results using different decoders will illustrate, the decoder needs to be designed carefully to ensure that the balance between feasibility and solution quality introduces a useful bias into the solution space, and does not eliminate high quality solutions within, or even close to, the feasibility boundary.

## 4   An Indirect Approach and the Three Decoders

This section starts with a description of an indirect GA approach whose main feature is a heuristic decoder that transforms the genotype of a string into its phenotype. After discussing the choice of encoding, three different decoders are presented and compared.

The first decision, when using a decoder based GA, has to be what the genotype of individuals should represent. Here the encodings are required to be of an indirect type, such that they represent an unconstrained problem and the decoder can build a schedule from it. Essentially, this leaves two possibilities in our case: The encoding can be either a permutation of the nurses to be scheduled or a permutation of the shifts to be covered. The former leads to strings of length $n$ for $n$ nurses (max 30) and the decoder would have to assign a shift pattern to each nurse. The latter gives strings of length equal to the number of grades times number of shifts, i.e. 3x14 = 42. In this case, the decoder would assign suitably qualified nurses to shifts. However, this approach leads to difficulties as the nurse preference cost $p_{ij}$ is given for full shift patterns, not for single shifts. Thus, we decided to use a permutation of the nurses. This has the further advantage that the multiple-choice constraint set (1) of the integer program formulation is implicitly fulfilled. Hence, this type of encoding requires a less sophisticated decoder.

Having decided to encode the genotypes as permutations of the nurses, a decoder that builds a schedule from this list has to be constructed. This 'schedule builder' needs to take into account those shifts that are still uncovered. Additional points to consider are the grades of nurses required, the types and qualifications of the nurses left to be scheduled and the preference cost $p_{ij}$ of a nurse working a particular shift pattern. Thus, a good schedule builder would construct feasible or near-feasible schedules, where most nurses work their preferred shift patterns. In the following, we present three decoders for this task, each with a different balance between feasibility and solution quality.

The first decoder is designed to consider only the feasibility of the schedule. It schedules one nurse at a time in such a way as to cover those days and nights with the highest number of uncovered shifts. The second is biased towards solution quality, but includes some aspects of feasibility by computing an overall score for each feasible pattern for the nurse currently being scheduled. The third is a more balanced decoder and combines elements of the first two. These are described in detail below.

The first decoder, referred to as 'Cover' decoder in the future, constructs solutions as follows. A nurse works $k$ shifts per week. Usually these are either all day or all night shifts (standard type). In some special cases, they are a fixed mixture of day and night shifts (special type). The first step in the 'Cover' decoder is to find the shift with the biggest lack of cover. This will decide whether the nurse will be scheduled on days or nights if the nurse is of the standard type. If there is a tie between, the nurse's general day / night preference will decide. We then proceed to find the $k$ day or $k$ night shifts with the highest undercover. If the nurse is of the special type, we directly find the *k* compatible shifts with the highest undercover, taking into account the number of day and night shifts worked by this particular nurse. The nurse is then scheduled to work the shift pattern that covers these *k* days or nights.

In order to ensure that high-grade nurses are not 'wasted' covering unnecessarily for nurses of lower grades, for nurses of grade *s*, only the shifts requiring grade *s* nurses are counted as long as there is a single uncovered shift for this grade. If all these are covered shifts of the next lower grade are considered and once these are filled those of the next lower grade. If there is more than one day or night with the same number of uncovered shifts, then the first one is chosen. For this purpose, the days are searched in Sunday to Saturday order. Due to the nature of this approach, nurses' requests (preference costs $p_{ij}$) cannot be taken into account by the decoder. However, they will influence decisions indirectly via the fitness function (see below for details), which decides the rank of an individual and the subsequent rank-based parent selection. Hence, the 'Cover' decoder can be summarised as:

1. Determine type of nurse.
2. Find shifts with corresponding largest amount of undercover.
3. Assign nurse to shift pattern that covers them.

The second decoder, called the 'Contribution' decoder, is designed to take account of the nurses' preferences. It therefore works with shift patterns rather than individual shifts. It also takes into account some of the covering constraints in that it gives preference to patterns that cover shifts that have not yet been allocated sufficient nurses to meet their total requirements. This is achieved by going through the entire set of feasible shift patterns for a nurse and assigning each one a score. The one with the highest (i.e. best) score is chosen. If there is more than one shift pattern with the best score, the first such shift pattern is chosen.

The score of a shift pattern is calculated as the weighted sum of the nurse's $p_{ij}$ value for that particular shift pattern and its contribution to the cover of all three grades. The latter is measured as a weighted sum of grade one, two and three uncovered shifts that would be covered if the nurse worked this shift pattern, i.e. the reduction in shortfall. Obviously, nurses can only contribute to uncovered demand of their own grade or below.

More precisely and using the same notation as before, the score $s_{ij}$ of shift pattern $j$ for nurse $i$ is calculated with the following parameters:

$d_{ks}$ = 1 if there are still nurses needed on day $k$ of grade $s$ otherwise $d_{ks}$ = 0.

$a_{jk}$ = 1 if shift pattern $j$ covers day $k$ otherwise $a_{jk}$ = 0.

$w_s$ is the weight of covering an uncovered shift of grade $s$.

$w_p$ is the weight of the nurse's $p_{ij}$ value for the shift pattern.

Finally, (100 - $p_{ij}$) must be used in the score, as higher $p_{ij}$ values are worse and the maximum for $p_{ij}$ is 100. Note that (-$w_p p_{ij}$) could also have been used, but would have led to some scores being negative. Thus, the scores are calculated as follows:

$$s_{ij} = w_p(100 - p_{ij}) + \sum_{s=1}^{3} w_s q_{is} \left( \sum_{k=1}^{14} a_{jk} d_{ks} \right)$$

The 'Contribution' decoder can be summarised as follows.

1. Cycle through all shift patterns of a nurse.
2. Assign each one a score based on covering uncovered shifts and preference cost.
3. Choose the shift pattern with the highest score.

The third 'Combined' decoder combines the bias towards feasibility of the 'Cover' decoder with features of the 'Contribution' decoder. It also calculates a score $s_{ij}$ for each shift pattern and assigns the shift pattern with the highest score to the nurse, breaking ties by choosing the first such shift pattern. However, in contrast to the 'Contribution' decoder, a shift pattern scores proportionally more points for covering a day or night that has a higher number of uncovered shifts. Hence, $d_{ks}$ is no longer binary but equal to the number of uncovered shifts of grade $s$ on day $k$. Otherwise using the same notation as before, the score $s_{ij}$ for nurse $i$ and shift pattern $j$ is calculated as before:

$$s_{ij} = w_p(100 - p_{ij}) + \sum_{s=1}^{3} w_s q_{is} \left( \sum_{k=1}^{14} a_{jk} d_{ks} \right)$$

Thus, the 'Combined' decoder can be summarised as follows.

1. Cycle through all shift patterns of a nurse.
2. Assign each one a score proportional to its contribution to uncovered shifts and preference cost.
3. Choose the shift pattern with the highest score.

Finally for all decoders, the fitness of completed solutions has to be calculated. Unfortunately, feasibility cannot be guaranteed, as otherwise an unlimited supply of nurses, respectively overtime, would be necessary. This is a problem-specific issue and cannot be changed. Therefore, we still need a penalty function approach. Since the chosen encoding automatically satisfies constraint set (1) of the integer programming formulation, we can use the following formula, where $w_{demand}$ is the penalty weight, to calculate the fitness of solutions. Note that the penalty is proportional to the number of uncovered shifts.

$$\sum_{i=1}^{n}\sum_{j=1}^{m} p_{ij} x_{ij} + w_{demand} \sum_{k=1}^{14}\sum_{s=1}^{p} \max\left[ R_{ks} - \sum_{i=1}^{n}\sum_{j=1}^{m} q_{is} a_{jk} x_{ij} ; 0 \right] \to \min!$$

## 5 Experiments

This section describes the computational experiments used to test the ideas outlined in the previous section. For all experiments, 52 real data sets as given to us by the hospital are available. Each data set consists of one week's requirements for all shifts and grade combinations and a list of nurses available together with their preference costs $p_{ij}$ and qualifications. The data was collected from three wards over a period of several months and covers a range of scheduling situations, e.g. some data instances have very few feasible solutions whilst others have multiple optima. Unless otherwise stated, to obtain statistically sound results all experiments were conducted as twenty runs over all 52 data sets. All experiments were started with the same set of random seeds, i.e. with the same initial populations. The platform for experiments was a Pentium II PC, which is similar to the hospital's equipment.

To make future reference easier, the following definitions apply for the measures used in the remainder of this paper: 'Feasibility' refers to the mean probability of finding a feasible solution averaged over all data instances and runs. 'Cost' refers to the sum of the best feasible solutions (or censored costs as described in the next paragraph) for all data sets averaged over all data sets. Thus, cost measures the unmet requests of the nurses, i.e. the lower the cost the better the performance of the algorithm.

Should the GA fail to find a feasible solution for a particular data set over all twenty runs, a censored cost of 100 is assigned instead. The value of 100 was used as this is more than ten times the average optimal solution value and significantly larger than the optimal solution to any of the 52 data instances. As a result of experiments described in Aickelin and Dowsland [9], the GA parameters and strategies in Table 1 were used for the experiments described in this paper. As can be seen from Table 1, the GA used is a standard generational implementation with fitness-rank-based roulette wheel selection and 10% elitism.

The purpose of the experiments described in this section was threefold. Firstly, to investigate the relationship between the heuristic decoder and the way in which it balances the dual objectives of feasibility and solution cost. Secondly, to compare the results produced by a number of different well-known permutation crossover operators; and thirdly to identify the most promising combination for further enhancement and possibly to identify some ways of achieving this.

The first objective was achieved by running the three decoders on their own and comparing the results. Once these results had been obtained, the 'Contribution' decoder was selected for further experimentation to investigate the sensitivity of the results to the way in which ties are broken. The results of the initial comparison are shown in the first three sections of Table 2. These results for the decoders without the GA are shown under the '100 random generations' label and are for 10,000 random nurse permutations, equivalent to approximately 100 generations with a GA or the length of a typical GA run. The results show that the decoders alone are not capable of solving the problem with on average less than 5% of runs finding a feasible solution.

It is interesting to note that on its own the 'Combined' decoder is superior to the other two and that between the remaining two the 'Contribution' decoder is the better choice for feasibility, but only marginally better in terms of overall cost than the 'Cover' decoder. When combined with the four variants of the GA all show a marked improvement in terms of both feasibility and cost. However, only the 'Combined' decoder has improved to an acceptable level, producing a large proportion of high quality feasible solutions. The 'Cover' decoder produces a reasonable proportion of feasible solutions, but they tend to be of relatively high cost. Although the GA obviously provides some pressure for cost improvement via the fitness function this is not sufficient to overcome the lack of influence of the $p_{ij}$ values in this decoder. The results of the 'Contribution' decoder are worse than those found by the 'Cover' decoder which is surprising as the decoder was designed to be more powerful. In this case, the GA does not seem to be able to guide the search towards permutations yielding feasible solutions. In view of its unexpectedly poor performance, this decoder was the subject of further experiments, aimed at 'tweaking' the heuristic in order to improve performance.

The most obvious modification is to change the decoder weights. It could be that the arbitrary choice (set intuitively to $w_1:w_2:w_3:w_P = 4:2:1:1$) conflicts with the GA. A set of experiments to examine this issue (not displayed) showed that the results were poor for any weight setting, although the ratio *8:2:1:1* performed slightly better than the others and was adopted in the following experiments.

A second possible modification is the way in which ties in the scores are broken. Shift patterns are searched in a given order with the first one encountered used in case of ties. In the above experiments, the patterns are ordered using the natural, or lexicographic, ordering, in which patterns involving the earlier days of the week appear first (referred to as 'Lexico'). This means that at the start of the schedule building when all shifts are equally uncovered there will be a tendency to fill the beginning of the week first. This is likely to have the effect of removing many feasible solutions from the solution space.

In view of this, four further orderings were investigated each designed to give a different balance between a fully randomised selection and bias guided by our knowledge of the problem. The first ('Rand Order') randomly shuffles the day and night shifts for each nurse separately, and then uses that order throughout the run. The second ('Biased') also uses a random ordering but starts with the day shifts with a 75% probability, reflecting the fact that the ratio between the covering requirements for days and nights is approximately 3:1. The third and fourth orderings order the patterns in increasing $p_{ij}$ order (using the lexicographic order to break ties). The third order ('Rand Cost') starts from a random point in this order, treating the ordering as a circular list, while the fourth ('Cheapest') starts searching at the start of this ordered list, i.e. with the cheapest pattern.

The results of experiments with these orderings are presented in graphical form in Figure 1. The random search orders achieve much better results than the two deterministic ones. Of the three orderings with random starting points, the 'Biased' is slightly better than the other two. Full results for this search order are displayed in Table 2 under the 'Contribution Biased' heading. The results have improved significantly, to the point where they are not far behind those of the 'Cover' decoder - but overall they are still disappointing, especially in view of the fact that the cost element $p_{ij}$ is included in this decoder.

One possible explanation could be that although both the 'Contribution' and 'Combined' decoders produce few feasible solutions on their own, the 'Combined' decoder accounts for both the number and magnitude of the under-covered shifts and hence the infeasible solutions tend to be closer to the boundary of the feasible region. This enables the genetic operators to produce more individuals that score well for both feasibility and cost. Overall, the results show that the choice of decoder is an important factor in the quality of solutions produced. Moreover, while it is not important that the decoder does well on an arbitrary set of permutations it is important that it should give rise to a solution space containing sufficient low cost solutions within or close to the feasibility boundary.

Table 2 also allows a comparison of four commonly used permutation crossover operators in conjunction with the different decoders. Ordered by the quality of results over all decoders, starting with the best, they are PMX (Goldberg and Lingle [24]), uniform order based crossover (Syswerda [25]), C1 crossover (Reeves [26]) and order-based crossover (Davis [27]). The rationale behind these experiments was that each type of crossover has different characteristics in terms of information passed to children regarding the order and position of genes. The percentage of genes which keep the same absolute position from one parent to a child are on average 33% for order based, 50% for C1 and uniform order based and 66% for PMX. Additionally for the order based, C1 and uniform order based crossovers the remaining genes keep the absolute order of the other parent. Hence, the results suggest that the higher the percentage of genes staying in their absolute position the better. Moreover, in the case of two operators with the same percentage, i.e. uniform order based crossover and C1 crossover, the more disruptive and hence flexible uniform crossover performs better. These results are not surprising, in that intuitively the good permutations will be those that schedule 'difficult' nurses first, leaving the easy ones until later. However, they do suggest that a crossover that is similar to uniform, i.e. not processing large blocks but keeping a higher proportion of genes in their absolute positions may result in further improvements.

## 6  Extensions

In the previous section, the 'Combined' decoder with PMX crossover was identified as the most promising variant for our nurse-scheduling problem. In this section we consider further modifications / enhancements leading to our final implementation. These are:

- Using the 'Biased' in place of the 'Lexico' ordering for the 'Combined' decoder.
- Fine-tuning the penalty weight used in the fitness function.
- Introducing a new crossover operator.
- Using a simple bound to improve the heuristic in the latter stages of the search.

The more detailed measure of undercover used in the score for the 'Combined' decoder means that the undesirable bias resulting from breaking ties with the 'Lexico' ordering will have less influence. Nevertheless, there will still be a tendency to allocate personnel to the beginning of the week. Therefore, the randomised orderings may give some improvement. A limited amount of experimentation was carried out to investigate this and indicated that the 'Biased' ordering is able to produce slightly better results. Therefore, this was adopted.

However, Even after this improvement, not all runs resulted in feasible solutions. Therefore, fine-tuning the weight of the preference cost $w_p$, was considered worthwhile. Figure 2 shows the outcome of a series of experiments for different values of $w_p$ indicated by the x-axis labels. As the graph shows, the results are sensitive to this parameter as it decides the balance between cost and feasibility. The behaviour for variations of $w_p$ is as would be expected. If it is set too low, then solutions are very likely to be feasible but are of high cost. If $w_p$ is set too high, solution quality rapidly drops due to the bias towards cheap but infeasible solutions. A value of $w_p = 0.5$ gives the best results, sacrificing only a small amount of feasibility for a good improvement in cost. However, the cost of solutions still leaves some room for improvement.

In an attempt to further improve results, a new type of crossover was introduced. This new crossover operator is inspired by the results found previously and combines the higher number of genes left in their absolute positions (like PMX) and re-ordering of blocks of information (like uniform permutation crossover). This new operator, called parameterised uniform order crossover or PUX for short, works in a similar way to uniform order based crossover as described in Syswerda [25]. However, when creating the binary template, the probability for a one will be equal to a parameter $p$, similar to standard parameterised uniform crossover [14]. For instance, if $p = 0.66$, then there will be a 66% chance for a one and hence a 66% chance that a gene will be in the same absolute position as in one of the parents.

Thus, PUX with $p = 0.66$ has an equal probability of keeping a gene in the same absolute position as PMX. However, PUX has an advantage. Whilst with PMX the remaining 33% of genes were positioned without any reference to the parents, PUX retains the absolute order of these as found in the second parent. In line with other uniform crossover operators, PUX is disruptive in a sense that it does not transmit large chunks of the parents. In order to find the best value for $p$, experiments were carried out shown in Table 2 under the 'Combined PUX' section. The value after each PUX label indicates the percentage used for $p$. (Note that $p = 50\%$ corresponds to the original uniform order based crossover). Particularly for the cost of solutions, results are further improved with an appropriate choice for $p$. For instance, for $p = 0.66$ feasibility is as high as for PMX, but solution cost is significantly lower. This suggests that our hypotheses about the necessary qualities of a successful crossover operator for our problem were right.

Another interesting observation is that the higher the value of $p$ the lower the feasibility of solutions. This indicates that a more disruptive crossover, i.e. $p = 0.5$, is more flexible and has the power to create feasible solutions when the other operators fail. On the other hand, solution cost is best for a medium setting of $p$, i.e. $p = 0.66$ or $p = 0.8$. This shows that for best results a careful balance has to be struck between flexibility and a parameter setting, which allows larger chunks to be passed on more frequently. In our case a setting of $p = 0.66$ is ideal.

The final enhancement of the indirect GA is based on the intelligent use of bounds. Herbert [28] concludes that results using a GA and a decoder can be improved with a new type of crossover operator. Herbert considers maximisation problems and uses a permutation based encoding with C1 crossover, which makes intelligent use of upper bounds, obtained from the fitness of partial strings. He argues that once a lower bound for the solution has been found only crossover points within that part of the string with an upper bound on fitness greater than the lower bound should be used. This is because the 'mistakes' in a particular solution must have happened before such a boundary point.

However, due to the difficulty of defining good upper bounds on partial fitness values of our strings, it is unclear how such a sophisticated operator would work in this case. Therefore, we propose a 'simple bound' based on a similar idea. When building a schedule from the genotypes it is easy to calculate the sum of the $p_{ij}$ costs so far. Once this sum exceeds a bound, set equal to the best feasible solution found so far, the schedule has 'gone wrong'. We could now employ backtracking to try to correct this. However, to guarantee that we improve the solution with the actual permutation of nurses at hand, a sophisticated algorithm of exponential time complexity would be necessary. This is outside the scope of this piece of research, but might be an idea for future work.

Instead, we propose a simpler approach. Once a feasible solution of cost $C^*$ has been found, we know that in the optimal solution no nurse $i$ can work a shift pattern $j$ with $p_{ij} > C^*$. We will use this as a rule when assigning shift patterns. In the early stages of optimisation this simple bound is of little use, as $C^* >> p_{ij}$. However, towards the end of the search when good feasible solutions have been found, the simple bound should prevent wasting time on dead-end solutions by making sure shift patterns with $p_{ij} \leq C^*$ are assigned.

Table 2 shows that the use of the simple bound slightly improved the cost of solutions whilst leaving the feasibility unchanged. This can be attributed to those runs where good solutions were found which were then further improved by forcing nurses onto even cheaper shift patterns. An additional benefit of using the simple bound was that average solution time was accelerated from around 15 seconds per single run to less than 10 seconds.

## 7  Summary of Results

A comparison of final results is shown in Figure 3, with details for the best algorithm in Figure 4. The simple 'Cover' decoder and 'Contribution' decoder produced mediocre results. Once both decoders were 'Combined' (label 'Combo'), the results rivalled those found with the most sophisticated direct GA including all enhancements [9] (label 'Direct'). However, once the PUX operator and the simple bounds were employed as well (label 'PUX'), the results were better than for any direct approach and within 2% of optimality.

A look at Figure 4 shows that 51 out of 52 data sets are solved to or near to optimality and a feasible solution is found for the remaining data set. The bars above the y-axis represent solution quality, with the black bars showing the number of optimal solutions, and the total bar height showing the number of solutions within three units of the optimal value. The value of three was chosen as it corresponds to the penalty cost of violating the least important level of requests in the original formulation. Thus, solutions this close to the optimum would certainly be acceptable to the hospital. The bars below the axis represent the number of times out of 20 that the run terminated without finding a single feasible solution. Hence the less the area below the axis and the more above, the better the performance.

Once the experiments had identified the best set of decoders and parameters, the final indirect GA was tested on a series of modifications to the problem specifications supplied by the hospital and on additional randomly generated data sets. In all cases, the modifications were easily added and the algorithm performed well. Moreover, results found for these new data sets were better than those for Dowsland's Tabu Search [8]. Thus, we feel our aim of creating a robust evolutionary algorithm for this problem has been achieved.

## 8   Conclusions

This paper presents an alternative algorithm for solving a nurse-scheduling problem in the form of a GA coupled with a decoding routine. In comparison to the previous 'direct' GA approach described in Aickelin and Dowsland [9] this has two advantages: Firstly, the GA solves an unconstrained problem leaving the constraint handling to the decoder that uses them to directly bias the search rather than in penalty functions alone. Secondly, all problem specific knowledge is held in the decoder routine, thus the algorithm can be quickly adapted to changes in problem specification. The overall results are better than those found by previous evolutionary approaches with a more flexible implementation than Tabu Search.

The nature of this type of decoder means that some of the conditions for an ideal transformation between the solution space and search space, as suggested by Palmer and Kershenbaum [11] cannot be satisfied. A comparison of three decoders showed that only when the decision making process struck the right balance between feasibility and quality good overall solutions were obtained. For problems that are more complex, one can imagine that setting these weights by hand will no longer be satisfactory. Consequently, some of our current research looks into possibilities of the algorithm finding its own weights intelligently with early results reported in Aickelin [29].

The role of the crossover operator was also shown significant. Again, this appears to be a question of balance – in this case between disrupting long sub-strings and inheriting absolute positions from the parents. The PUX operator was introduced to facilitate such a balance and the results of experiments with this operator suggested that more disruptive operators helped feasibility, but that too much disruption affected solution quality.

The quality of the final results, together with the fact that the indirect GA approach proved to be more flexible and robust than Tabu Search, makes the indirect GA a good choice to solve the nurse-scheduling problem. Its central idea of changing the problem into a permutation problem and then building solutions with a separate decoder can be applied to all constrained scheduling problems. Thus given this success, experiments with similar approaches on other difficult problems, in scheduling and other application areas is an interesting area for further research.

| Parameter / Strategy | Setting |
|---|---|
| Population Size | 100 |
| Population Type | Generational |
| Initialisation | Random |
| Selection | Proportional to Fitness-Rank |
| Crossover | Two parents, Order-Based |
| Bit Swap Mutation Probability | 1.5% |
| Replacement Strategy | Keep 10% Best |
| Stopping Criteria | No improvement for 30 generations |
| Penalty Weight | 20 |

Table1: Parameters and strategies used for the indirect GA and nurse scheduling following [9].

| Decoder | Type of indirect GA | Time [s] | Feasibility | Cost |
|---|---|---|---|---|
| Cover | 100 random generations | 29.2 | 2.7% | 191.3 |
| | C1 Crossover | 19.7 | 42.1% | 68.7 |
| | Order Crossover | 17.4 | 39.2% | 69.7 |
| | Uni Crossover | 17.4 | 45.4% | 63.3 |
| | PMX Crossover | 18.5 | 46.4% | 61.7 |
| Contribution | 100 random generations | 27.3 | 3.2% | 190.2 |
| | C1 Crossover | 18.9 | 13.9% | 154.2 |
| | Order Crossover | 17.2 | 12.7% | 155.9 |
| | Uni Crossover | 16.5 | 14.9% | 151.5 |
| | PMX Crossover | 17.1 | 16.6% | 144.9 |
| Combined | 100 random generations | 25.2 | 5.2% | 171.0 |
| | C1 Crossover | 14.0 | 92.3% | 15.0 |
| | Order Crossover | 14.4 | 88.4% | 19.0 |
| | Uni Crossover | 15.0 | 97.4% | 14.1 |
| | PMX Crossover | 15.3 | 96.6% | 14.3 |
| Contribution Biased | 100 random generations | 27.1 | 4.2% | 142.1 |
| | C1 Crossover | 19.9 | 39.2% | 71.2 |
| | Order Crossover | 19.4 | 37.3% | 74.5 |
| | Uni Crossover | 19.7 | 42.1% | 68.7 |
| | PMX Crossover | 18.9 | 45.6% | 67.5 |
| Combined Biased PUX | PUX 50% | 15.0 | 97.4% | 14.1 |
| | PUX 66% | 14.6 | 96.6% | 10.0 |
| | PUX 80% | 13.9 | 96.0% | 10.2 |
| | PUX 90% | 13.8 | 95.4% | 14.3 |
| | PUX 66% and simple bound | 9.3 | 96.6% | 9.4 |
| Summary | Indirect GA | 9.3 | 96.6% | 9.4 |
| | Direct GA [9] | 14.9 | 90.9% | 10.8 |
| | TABU [8] | 30 | 100% | 8.8 |
| | IP [7] | >6000 | 100% | 8.7 |

Table 2: Comparison of Results. Lower cost, respectively higher feasibility is better.

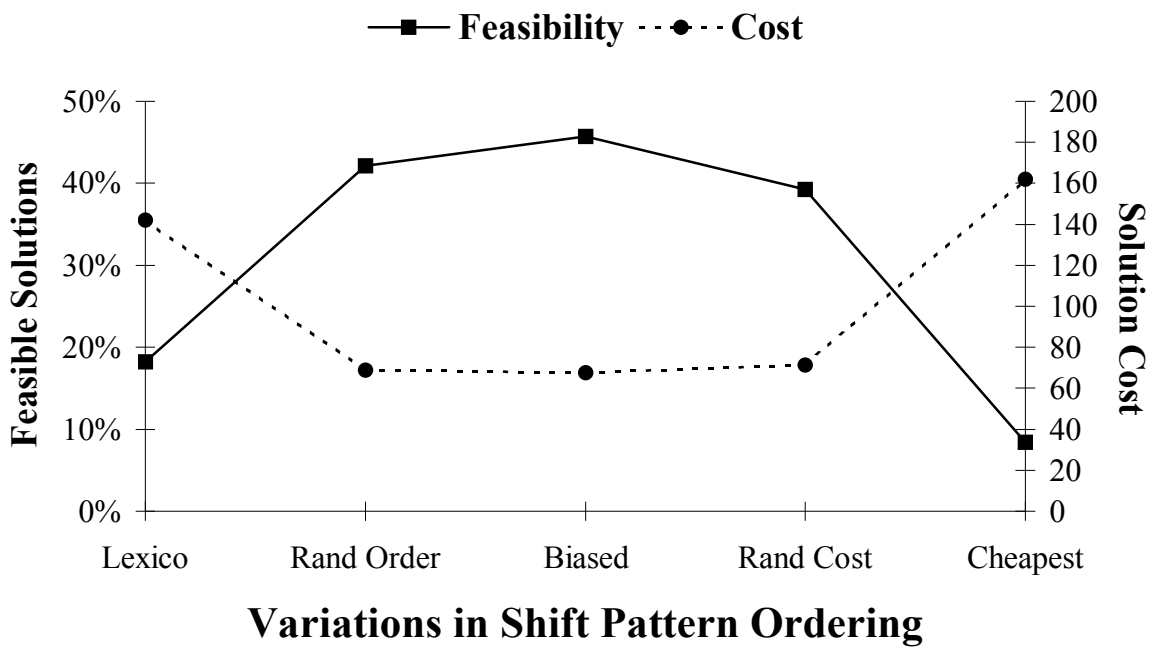

Figure 1: Effect of different search orders for the 'Contribution' decoder. Lower cost, respectively higher feasibility is better.

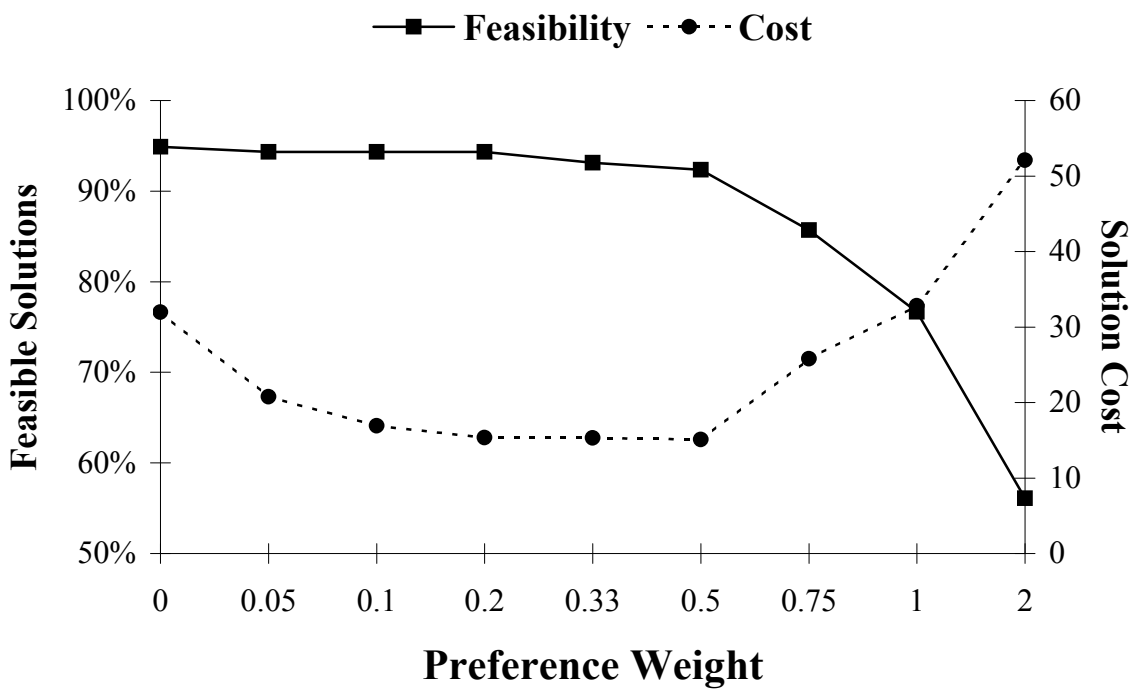

Figure 2: Variations of preference weight for the 'Combined' decoder and its effect on solution quality. Lower cost, respectively higher feasibility is better. Note that the x-axis is not to scale.

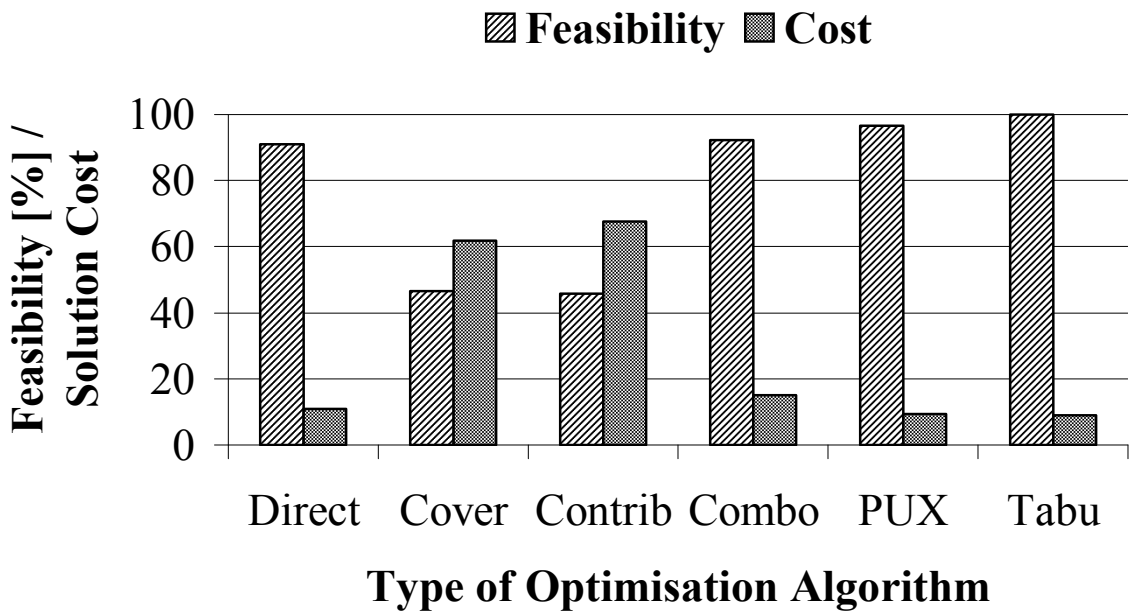

Figure 3: Comparison of results of various heuristics approaches to the nurse-scheduling problem. Lower cost, respectively higher feasibility is better.

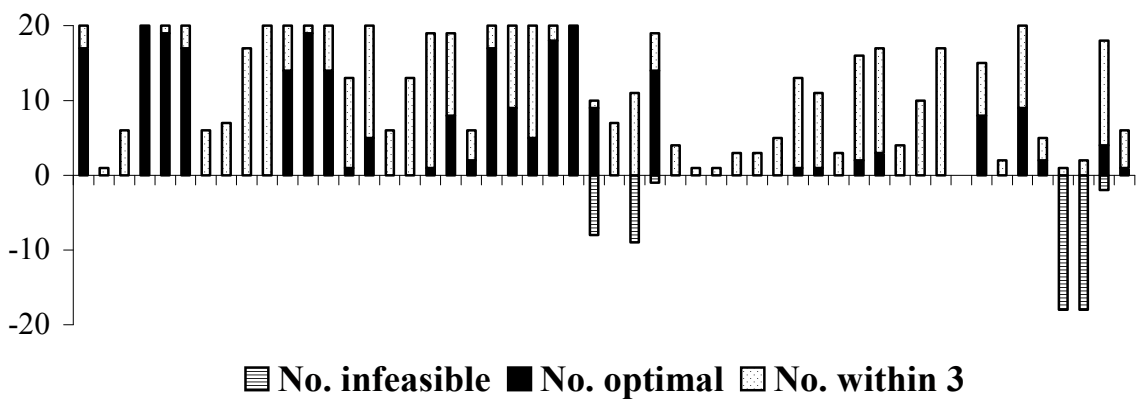

Figure 4: Detailed results for all 52 data sets using the enhanced indirect GA. The bars show the number of infeasible, good feasible and optimal solutions out of 20.